\documentclass[sigconf, nonacm]{acmart}

\usepackage{multirow}
\usepackage{subfigure}
\usepackage[linesnumbered]{algorithm2e}
\usepackage[noend]{algpseudocode}
\usepackage{enumitem}

\usepackage{graphicx}

\begin{document}

\title{Scaling Knowledge Graph Embedding Models}
%%
%% The "author" command and its associated commands are used to define the authors and their affiliations.
\author{Nasrullah Sheikh}
\orcid{0000-0001-7194-9385}
\affiliation{%
  \institution{IBM Research Almaden}
  \city{San Jose}
  \state{California}
  \country{US}
}
\email{nasrullah.sheikh@ibm.com}

\author{Xiao Qin}
\affiliation{%
  \institution{IBM Research Almaden}
  \city{San Jose}
  \state{California}
  \country{US}
}
\email{xiao.qin@ibm.com}

\author{Berthold  Reinwald}
\affiliation{%
  \institution{IBM Research Almaden}
  \city{San Jose}
  \state{California}
  \country{US}
}
\email{reinwald@us.ibm.com}

\author{Chuan Lei}
\affiliation{%
  \institution{Instacart}
  \city{San Francisco}
  \state{California}
  \country{US}
}
\email{chuan.lei@instacart.com}

\begin{abstract}
Developing scalable solutions for training Graph Neural Networks (GNNs) for link prediction tasks is challenging due to the high data dependencies which entail high computational cost and huge memory footprint. We propose a new method for scaling training of knowledge graph embedding models for link prediction to address these challenges. Towards this end, we propose the following algorithmic strategies: self-sufficient partitions, constraint-based negative sampling, and edge mini-batch training. Both, partitioning strategy and constraint-based negative sampling, avoid cross partition data transfer during training. In our experimental evaluation, we show that our scaling solution for GNN-based knowledge graph embedding models achieves a 16x speed up on benchmark datasets while maintaining a comparable model performance as non-distributed methods on standard metrics.
\end{abstract}

\maketitle

\begingroup
\renewcommand\thefootnote{}\footnote{\noindent
This work is licensed under the Creative Commons BY-NC-ND 4.0 International License. Visit \url{https://creativecommons.org/licenses/by-nc-nd/4.0/} to view a copy of this license. For any use beyond those covered by this license, obtain permission by emailing. \\
}
\endgroup

%%% VLDB block end %%%

\section{Introduction}

Graphs have been widely used to model and manage relational data~\cite{graphgen}. 
Knowledge graphs (KG), as a prime example of graphs, model real-world objects, events, and concepts as well as various relations among them. 
Representation learning over large-scale knowledge graphs has been emerging as a pivotal tool to derive insights from graph structured data powering a wide range of applications such as data integration~\cite{benchmark, kgalign, avrgcn} and question answering~\cite{qakb, bordes2015qa}.

KG embedding methods~\cite{kgs1, transe, kgcomp,rgcn, relgnn, kbgan, zhang-kge} capture the attributes of entities and structures of relations in KGs, and project them into a lower dimensional vector space for use in various downstream tasks such as vertex classification~\cite{rgcn} and link prediction~\cite{rgcn, relatt, relgnn, distmult,transe}. 
Traditional KG embedding methods learn various structures pattern between the entities such as \textit{symmetric}, \textit{anti-symmetric} and \textit{inverse} relations. They mainly focus on the scoring aspect of the problem, which is to predict the legitimacy between two entities and a particular relation type.
Recently, message passing-based graph neural networks (GNNs) have been adopted to enhance the expressivity of the entity\footnote{Entity and vertex terms are used interchangeably.} embeddings~\cite{rgcn}. GNNs capture the topological features of the entities such as shapes of the neighborhood sub-graphs which are overlooked by the traditional KG embedding methods. 
\begin{figure}
  \centering
  \includegraphics[width=\linewidth]{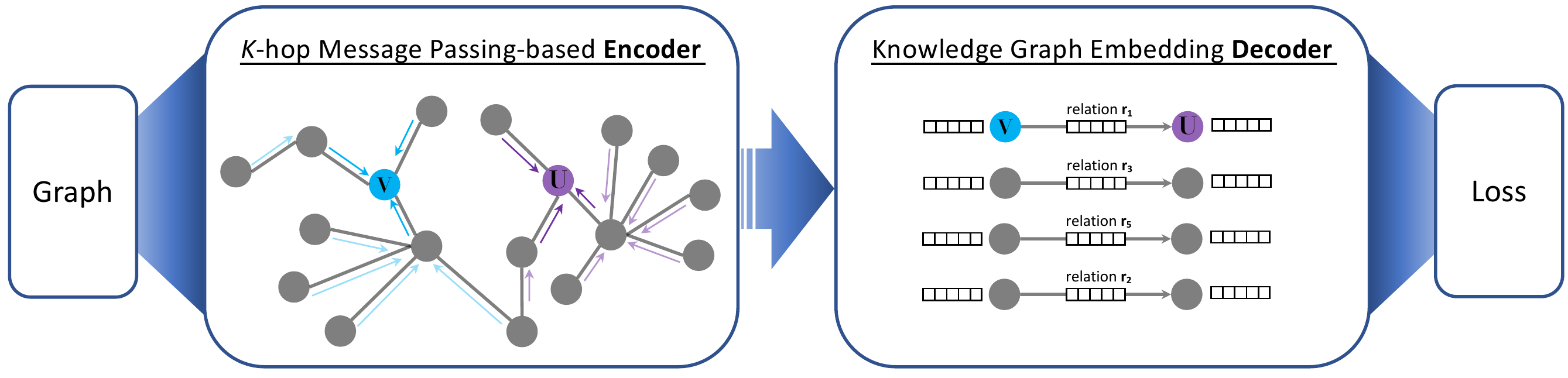}
  \caption{A graph neural network-based knowledge graph embedding network architecture.}
  \label{fig:encode-decode}
\end{figure}
A generic GNN-based KG embedding neural network (Figure~\ref{fig:encode-decode}) demonstrates an encoder-decoder architecture. The encoder generates entity embeddings using a GNN by aggregating neighbors' information, and the decoder learns the relations between entities using a traditional KG embedding method. Such architecture enjoys the benefits from both methods promising to generate expressive entity and relation embeddings. However, it learns better embedding but at the cost of increased model complexity in terms of number of trainable parameters. For example, TransE\cite{transe} has 1.5 million parameters where as RGCN\cite{rgcn} has 3.3 million parameters over FB15k-237 dataset with an embedding size of 100. The increasing number of trainable parameters leads to an increase in the training time. 

Besides the model complexity, the size of the modern input KGs has also grown exponentially.
The Facebook~\cite{fbgraph} graph has billions of vertices and trillions of edges, and Freebase~\cite{freebase} has millions of entities and billions of edges. Iterative training on these large graphs may not be feasible on single node systems due to its high computational cost and its significantly high memory requirements caused by the data dependencies.

\begin{figure}[h]
  \centering
  \includegraphics[width=\linewidth]{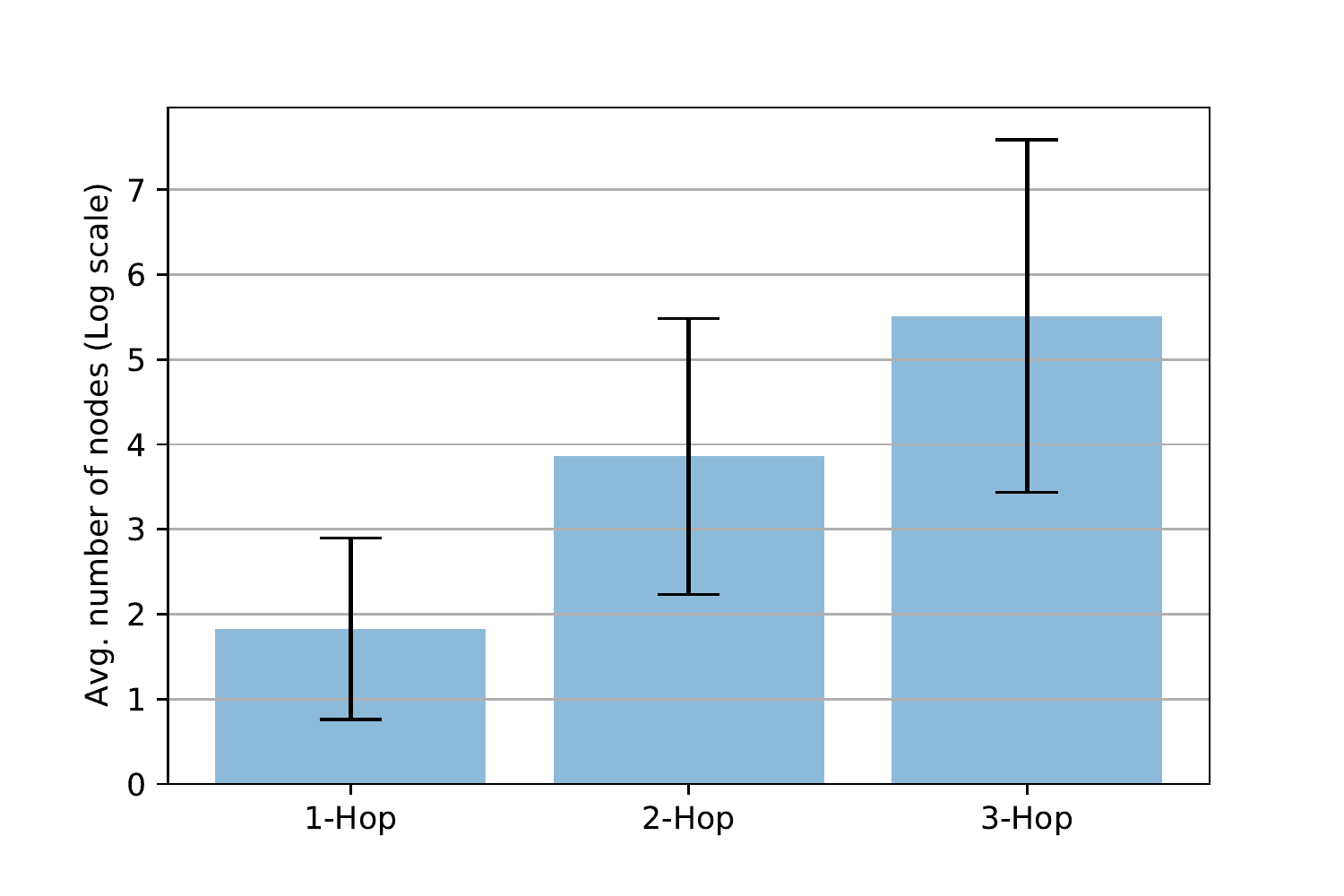}
  \caption{Average number of vertices required to compute the embedding of a vertex in ogb-citation2 dataset.}
  \label{fig:hops}
\end{figure}

Various distributed training frameworks~\cite{dglke,pbg,graphvite} have been proposed to scale KG embedding methods. However, these frameworks apply to the traditional models (various types of decoder in Figure~\ref{fig:encode-decode}) with mutually independent training triplets~\cite{distmult, transe}. The input data can be partitioned easily, and the models are subsequently trained in parallel. These frameworks cannot be used for training GNN-based KG embedding models~\cite{rgcn,relgnn} due to the inherent dependencies on the neighborhood information (usually beyond $n$-hop with $n \geq 2$). Figure~\ref{fig:hops} shows that with larger and deeper neighborhoods, the average number of vertices required to compute an embedding rise significantly, which consequently leads to an increase in the number of model parameters. Thus, it requires larger computation and memory requirements.
We further observe that the skewed distribution of vertex degrees in enterprise knowledge graphs leads to vertex dependencies up to tens of thousands of vertices. These dependencies make scaling GNN-based KG embedding models extremely challenging.

Several distributed GNN training frameworks have been proposed primarily for the vertex classification task~\cite{distdgl, neugraph}. Graph partitioning followed by distributed training are commonly explored by these solutions. While a simple partitioning strategy is to partition the graph using either vertex-cut or edge-cut-based methods, and access required dependent vertices in other partitions remotely during training. However, with increase of the number of GNN layers, i.e., the number of hops, the amount of exchanges of neighborhood information across partitions will incur significant communication overhead. It is in contrast to distributed neural network training on non-graph structured datasets such as images, which only incurs communication overhead due to gradients sharing. The exchange of neighborhood information is the main bottleneck in scaling GNN training. The challenge is to generate optimized graph partitions that reduce the required exchange of neighborhood information. Moreover, partitions generated from larger graphs are of considerable size and cannot fit in the smaller memory such as GPU memory. The another challenge is to train on these considerably larger partitions, while still taking advantage of GPU. DistDGL~\cite{distdgl} introduces an edge-cut-based partitioning method using METIS~\cite{metis}, and employs a mini-batch training approach for vertex classification. Edge-cut-based methods produce partitions with edge replication in multiple partitions, which may lead to skewed partition sizes. The larger partitions will be the stragglers in the training process. We observe that partitions produced by METIS followed by neighborhood expansion for link prediction are approximately 33\% larger than the partitions produced by vertex-cut based methods~\cite{kahip}, which increases the training time by approximately 21\%. Thus, this approach is not optimal for the link prediction task.

In this paper, we propose a distributed training approach using GNN-based knowledge graph embedding models for link prediction. We introduce a vertex-cut method to partition the graph and then expand the partition to include $n$-hop neighbors, where $n$ is determined by the number of convolutional layers of the GNN model. The partitions produced are self-sufficient, thus do not require any exchange of neighborhood information during distributed training, but at the expense of data replication and redundant computation. Moreover, we generate negative samples within the partition, thus further reducing the communication overhead. Using data parallel approach, we train the model in a cluster where each trainer process trains on a partition using edge mini-batch training strategy.

Our main contributions are as follows.

\begin{itemize}[leftmargin=*]
\item To the best of our knowledge, we propose the first architecture for distributed GNN-based knowledge graph embedding model training for link prediction. We also introduce edge mini-batch training which allows us to train on large partitions. 
\item We employ a vertex-cut-based partitioning strategy that partitions the graph into sets of disjoint edges, which we then expand to self-contained graph partitions by replicating $n$-hop dependent vertices and edges required for message passing. 

\item We exploit the locally closed world assumption~\cite{complex, transe} and employ a constraint based negative sampling strategy to sample negative samples. The negative samples are drawn from within the partition to avoid communication overhead. The approach allows negative sampler to generate negative samples locally within the partition, thus avoiding communication cost of fetching the negative samples from other partitions. 
\item We experimentally evaluated the performance of our proposed system on two public datasets. Our approach achieves a speedup of 16x with 8 trainers without any loss on measured metrics.
\end{itemize}
\section{Background}
\label{sec:bg}

\subsection{Knowledge Graph Neural Networks}
\label{subsec:gnn}

Knowledge graph convolutional network based graph embedding models such as~\cite{rgcn, relgnn} use a message passing framework and apply a relation-specific transformation on the information from neighbors to obtain the embedding of a head vertex. A general message passing framework for knowledge graph embedding can be expressed as:
\begin{equation}
\label{eq:mpass}
    h_s^{\prime} = \sigma \left( \textsl{Agg}_{(r,t) \in \mathcal{N}_s} f(h_s, r, h_t)\right)
\end{equation}
where $f$ is the relation-specific transformation function applied on the information from neighbors, $\textsl{Agg}$ is an aggregator function such as \textit{SUM}, \textit{MEAN} that is applied on the received information from all neighbors, $N_s$ defines the  neighbors of node $s$, $r$ is the relation between $s$ and $t$ nodes, $\sigma$ is the activation function, and $h_s^\prime$ is the new hidden representation of entity $s$. To prevent overfitting, the paper~\cite{rgcn} introduced two techniques for regularisation: 1) Basis decomposition and 2) Block diagonal decomposition. In case of basis decomposition, the weight  matrix $W_r$ is obtained as a linear combination of smaller basis matrices $V_b \in \mathbb{R}^{d \times d}$ as defined below:
\begin{equation}
    W_r = \sum_{b=1}^{B}a_{rb}V_b
\end{equation}
where $a_{rb}$ are coefficients that depend on $r$.

A knowledge graph embedding model is trained using a negative sampling approach. For each positive triplet $\tau \in T^+$, a set of negative samples are generated by either corrupting $s$ or $t$ which produces a set of negative triplets $T^-$. Given the set of positive and negative triplets $T=T^+ \bigcup T^-$,  the model is optimized on cross-entropy loss to learn entity and relation embeddings.
\begin{equation}
    \label{eq:loss}
    \mathcal{L} = \frac{1}{|T|} \sum_{\tau \in T} y~\textsl{log}~l\big(g(
    \tau)\big) + (1-y)\textsl{log}\big(1- l(g(\tau))\big)
\end{equation}
where $\tau$ is a training example $(s,r,t)$; $l$ is the logistic \textit{sigmoid} function; $y$ is 1 or 0 for  positive or negative triplets, respectively; and $g$ is a scoring function such as DistMult~\cite{distmult} as shown below:
\begin{equation}
    \label{eq:scoringfun}
    g(s,r,t) = h_s^T\,\mathbf{M}_r\,h_t
\end{equation}
where $\mathbf{M}_r$ is a diagonal matrix associated with each relation $r$.

\subsection{Data Parallel Distributed Training}

In data parallel distributed training, a model is replicated to each compute node, and each replicated model works on a partition of the data independently. The distributed training must have mathematical equivalence to the non-distributed training, i.e., the results produced by distributed training should be same as in non-distributed training. There are two approaches to make replicated models consistent after each iteration: gradient sharing or parameter sharing. With gradient sharing, each replicated model is updated by exchanging and averaging its gradients before the optimizer step. This enables consistency across all replicated models. In case of parameter sharing, each replicated model shares its parameters and the model parameters are averaged. This step is performed after the optimizer step. Parameter averaging does not satisfy the mathematical equivalence requirements, which leads to a loss in model accuracy~\cite{pydist}. Moreover, parameter sharing under utilizes the resources as the optimizer step typically being the hard synchronization point, when the workers remain idle during the communication.

The gradients can be shared using the classical parameter server paradigm or AllReduce. In parameter server paradigm, a machine is responsible for aggregating the gradients from worker machines,  and then sending back the averaged gradients to the workers. This approach requires high bandwidth for gradient sharing, and the gradient computation and communication are non-overlapping operations.  In AllReduce gradient sharing paradigm,
each compute node performs the aggregation  on a subset of parameters. Thus, reducing the required communication bandwidth. Moreover, gradient computation and communication operations are overlapping. AllReduce is supported by several communication libraries such as NCCL~\cite{nccl}, MPI~\cite{mpi}, and GLOO~\cite{gloo}. 
PyTorch~\cite{pytorch} provides an API, \textit{DistributedDataParallel} for distributed data parallel training, which uses an AllReduce communication library for gradient sharing. Shen et al. has provided a detailed analysis of data parallel training in PyTorch~\cite{pydist}. Various libraries such as RaySGD\cite{ray} and Horovod\cite{horovod} have been developed that provide wrappers around DistributedDataParallel. Besides ease of use, the libraries improve performance and allow efficient utilization of resources.

\section{Distributed Knowledge Graph Training}
In this section, we describe our distributed learning process of GNN-based KG embedding models on a cluster of compute nodes with multi CPUs/GPUs.
\begin{figure}[t]
  \centering
  \includegraphics[width=\linewidth]{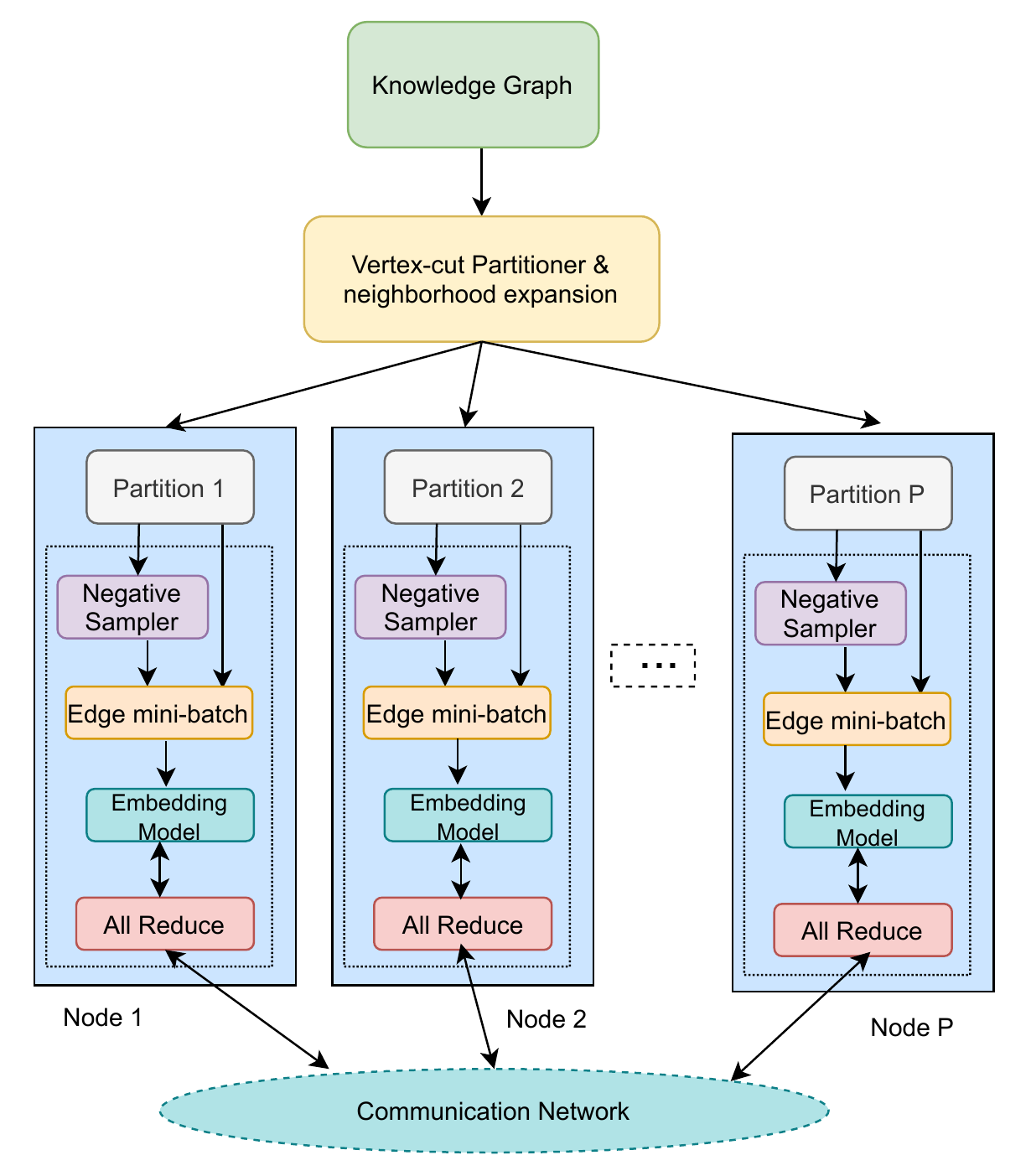}
  \caption{Architecture of our distributed approach for link prediction.}
  \label{fig:arch}
\end{figure}

\subsection{System Overview}

The architecture\footnote{For the purpose of the overview, we assume one compute node runs only one worker.} of our proposed approach is shown in Figure~\ref{fig:arch}.  We choose the \textit{AllReduce} paradigm of gradient sharing over the parameter server approach because it incurs less communication cost, and computation and communication is being overlapped during gradient sharing. Our proposed architecture is designed to run on a distributed CPU/GPU cluster. Each compute node (CPU/GPU) in a cluster runs a replica of the model and is responsible for training on a partition of data using synchronous gradient descent (SGD). Each training process computes the gradients of the model on an edge mini-batch, shares and averages the gradients, and updates the local model. Specifically, our distributed KG embedding learning involves the following steps:

\begin{enumerate}[leftmargin=*]
    \item Partition the graph into $P$ disjoint subsets, and then expand each partition to include $n$-hops of neighbors of each vertex in the partition, where $n$ is determined by the number of graph convolutional layers in the embedding model. The number of partitions is equal to the number of compute nodes available. We refer to compute node as a processing unit (CPU/GPU). Graph partitions along with the required features of vertices are assigned to a compute nodes.
    \item One training process/worker is launched per compute node. During each epoch, each training process samples $s$ negative samples for each positive sample in its partition. The number of training examples in a partition is $p \times (s+1)$, where $p$ is the number of positive samples in a partition.
    \item Each training process implements edge mini batching for training. A batch of $b$ edges (positive and negative) in a partition is sampled. The edge mini-batch ensures that the embedding of all entities required for scoring the edges in the edge mini-batch is computed. 
    \item After the formation of an edge mini-batch, a computational graph is generated for message passing in the graph convolutional layers. We obtain the loss, and compute the gradients. The gradients are shared using AllReduce, and the model is optimized based on the averaged gradients. Our proposed approach can be applied to any graph embedding model which uses a message passing approach for graph convolution. 
\end{enumerate}

The enumerated steps in the overview are described in detail in the following subsections.

\subsection{Graph Partitioning}
\label{subsec:graphpart}

Partitioning the input graph is an important preprocessing step in distributed training. The quality of partitions have a direct impact on the learned model quality and on scale. We apply a two-phase approach by first partitioning the graph and then performing the neighborhood expansion to make the partitions self sufficient. 

\begin{figure}[!h]
  \centering
  \includegraphics[width=\linewidth]{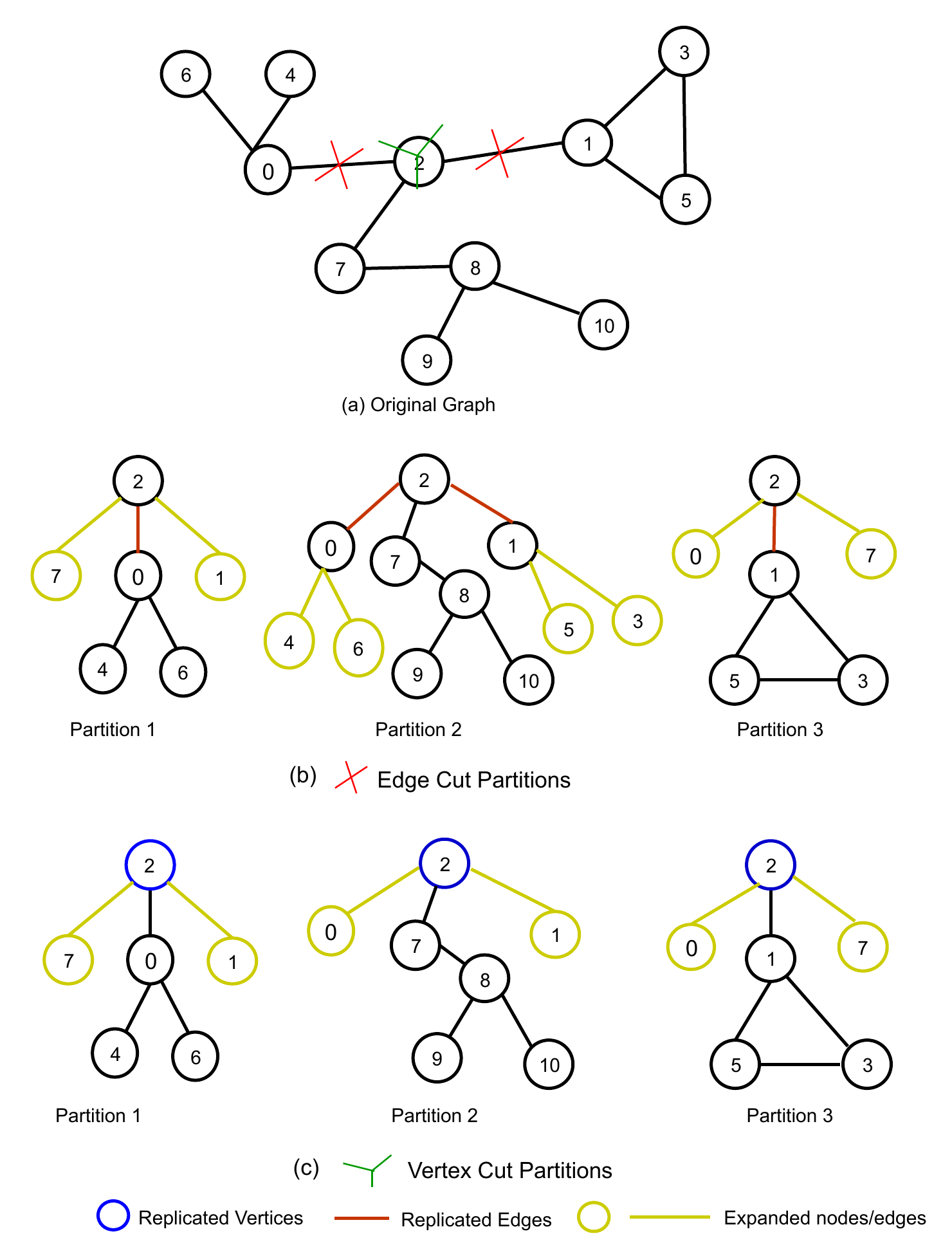}
  \caption{Graph partitioning: edge cut and vertex cut partitions along with their neighborhood expansion.}
  \label{fig:gpart}
\end{figure}

\subsubsection{Partitioning} 
Using edge-cut partitioning methods such as Metis\cite{metis}, some positive edges are replicated in multiple partitions as shown in Fig~\ref{fig:gpart}(b) (e.g., Edges $(0,2), (1,2)$ are present in all partitions). Hence, the training on the replicated edges is repeated in multiple partitions which incurs additional computational cost, and may also negatively impact the learning process. Moreover, edge-cut partitioning is shown to be ineffective in balancing the workload of large real-world graphs~\cite{dist_edge_partitioning, graph_part_NE}. This load imbalance leads to a substantial stalling of work which increases the overall training time.

Vertex cut partitioning~\cite{schlag2018scalable, dist_edge_partitioning, graph_part_NE} divides the edges into disjoint partitions and produces balanced partitions by minimizing the vertex replication. We refer to edges in a partition as \textit{core edges}, the vertices where the graph is partitioned as \textit{replicated-vertex}, and other vertices as \textit{core-vertex}. The set of \textit{core edges} form the positive edges for training.  

The disjoint partitions produced by vertex-cut partitions are more suited for the problem of link-prediction because the produced partitions are balanced and also the neighborhood expansion does not lead to graph explosion (discussed in detail in ~\ref{subsec:ne}).

\subsubsection{Neighborhood Expansion}
\label{subsec:ne}
Link prediction requires an updated embedding of vertices of an edge to calculate the score to determine the validity of an edge. As discussed in the above section~\ref{subsec:gnn}, to compute an embedding of a vertex, a $n$ layer GNN requires to have the features from the $n$-hop neighbors. Due to partitioning, some edges will have partial neighborhood information available within the partition, and the other required information could be present in different partitions. We call these edges as \textit{boundary-edges}. One possible way is to fetch this information during the training. But this will incur the communication cost, thus incurring delay in the training. We propose to make the partitions independent by including the missing partial neighborhood information of \textit{boundary-edges}. We call this process as \textit{Neighborhood Expansion}, and it is done after creating partitions. This eliminates the communication cost of fetching data from other partitions, but at the expense of the increased size of the partitions. We refer to these added vertices and edges as \textit{support-vertices} and \textit{support-edges} respectively.  The neighborhood expansion of the graph is shown in Figures~\ref{fig:gpart}(b) and (c). It can be seen from Figure~\ref{fig:gpart}(b) that edge-cut partitions explode during Neighborhood Expansion, thus limiting its usability.
 
\subsection{Training}
Each compute node will have a replica of the graph embedding model and will work on a single partition of data. The partition assigned to a compute node remains fixed during the entire training process. We generate a set of negative edges, and the combined set of negative edges and \textit{core edges} form the set of training edges. We employ a edge mini-batch training approach to train the KG embedding methods. The distributed training process is given in Algorithm~\ref{algo:train}.

\subsubsection{Negative Sampling}
\label{subsec:negsamp}
In KG embedding methods, the negative samplers generally exploit the closed world hypothesis which considers any triplet not explicitly present in the graph as a negative triplet. Most of the negative triplets generated by this result are easy negative examples~\cite{neganalysis}. The error gradient from these samples are very small, hence does not help in learning. The negative samples space $O(N^2)$ is far larger than positive samples space, thus more prone to generating easy negative samples. 

We propose a constraint-based negative sampling approach. Our proposed approach considers the core edges in a partition as the positive triplets and each partition is independent of other partitions. We employ a constraint that generates the negative samples from the core vertices of the partition based on the local world hypothesis. This constraint provides two advantages, 1) the embeddings of entities in negative samples are not stale, and 2) it avoids the communication cost of querying other partitions and fetching data. This also reduces the sample space of negative samples as $N << N_i$, where $N_i$ number of vertices in the $i$-th partition and $N$ is the total number of vertices in a KG, respectively. This helps in minimizing the problem of generating easy negative samples.

\subsubsection{Edge Mini-Batch}
\begin{figure}[t]
  \centering
  \includegraphics[scale=0.9]{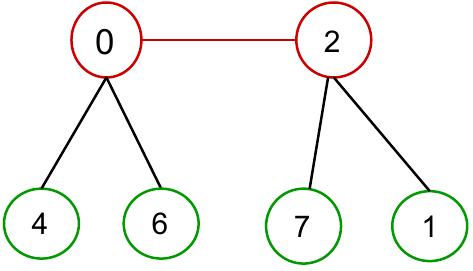}
  \caption{Edge Mini-batch}
  \label{fig:edgebatch}
\end{figure}

GNN training on a large dataset for vertex classification is done by mini-batching. In mini-batching, a set of vertices are randomly selected, and a computational sub-graph is sampled for training to obtain the embeddings of the selected vertices. Using a vertex sampling strategy for link prediction is not trivial as it does not guarantee that both vertices of an edge are in the sample. For this reason, we are proposing to use \textit{edge mini-batching} for link prediction. In \textit{edge mini-batching}, a batch of edges are sampled, and the vertices in the batch form a vertex set. Then, a computational graph for message passing is created which captures the $n$-hop dependencies of the sampled edge batch. The embeddings are learned for vertices that are in the vertex set. Figure~\ref{fig:edgebatch} shows a $1$-hop computational graph for an edge $(0,2)$, and message passing is done on this graph to learn the embeddings of vertex $0$, and $2$.

\subsubsection{Model Training}
For each edge mini-batch training, we generate a computational graph from the partitioned data using the vertices in the graph. This computational graph is used for embedding generation for the vertices in the the edge mini-batch using the graph convolutional layers of the KG embedding model. This constitutes the forward pass of the training. In the next step, a loss is calculated for the set of edges in the mini-batch which subsequently generates gradients. These gradients are shared among the training processes using AllReduce communication primitive. Once the gradients are shared and averaged, the training processes update their local model.

\SetKwComment{Comment}{/* }{ */}
\RestyleAlgo{ruled}
\begin{algorithm}
\caption{Training Process on each compute node}\label{algo:train}
\KwIn{GNNmodel, optimizer, features, gPartition, epochs}
$epoch \gets 1$\\
\While{$epoch \leq epochs$}{
    $negativeEdges \gets negativeSampler(gPartition)$ \\
    \While{$batch \in batches(negativeEdges, gPartition)$}{
    $cgraph \gets getComputeGraph(batch, gPartition)$\\
    $embedding \gets GNNmodel(features, cgraph)$ \\
    $loss \gets loss(embedding, batch)$ \\
    $backward(loss)$
    \Comment*[r]{\scriptsize{Gradients are computed \& shared}}
    $optimizer.step()$ \Comment*[r]{\scriptsize{The model is updated}}
    }
  $epoch \gets epoch + 1$\\
  }
\end{algorithm}

\section{Experimental Evaluation}

\subsection{System Setup}
We ran our experiments on a cluster of 4 machines. Each node has two Intel Xeon 6138 CPUs @ 2.00 GHz (80 virtual cores), 726 GB DDR4 DRAM @ 2666 MT/s, 40 Gb Ethernet, 2 P100 GPUs, and running CentOS Linux 7.9.  

We use PyTorch Geometric 1.7.2~\cite{pygeom} as graph embedding framework and PyTorch 1.9.0~\cite{pytorch} as the deep learning backend framework. We use PyTorch \textit{DistributedDataParallel} for distributed training, and Gloo~\cite{gloo} as the backend for collective communication operations for GPUs and \textit{AllReduce} for gradient sharing.

\subsection{Evaluation Metrics}
The performance of link prediction in knowledge graphs is evaluated using MRR (Mean Reciprocal Rank) and Hits@$k$. For each test triplet $p$, a set of corrupted triplets is generated by either corrupting $h$(ead) or $t$(ail). These corrupted triplets are filtered by removing the triplets which occur in the training data. This procedure is referred to as "filtered setting". Let  $T$ be the set containing a positive triplet $p$ and its respective corrupted triplets arranged in descending order of their scores, and $r_p$ is the rank position of $p$ in $T$. MRR in Equation~\ref{eq:mrr} is the average reciprocal rank of the positive triplets, and Hits@$k$ in Equation~\ref{eq:hits} is the fraction of positive triplets that appear in first $k$ highest ranked triplets, where $N$ is the total number of positive triplets in the test set.

\begin{equation}
\label{eq:mrr}
    MRR = \frac{1}{N} \sum_{p=1}^N \frac{1}{r_p}
\end{equation}

\begin{equation}
\label{eq:hits}
    Hits@k = \frac{|\{p|r_p <= k\}|}{N}
\end{equation}

\subsection{Datasets}
We experimented with two public benchmark datasets used for link prediction in KGs.  The dataset statistics are summarized in Table~\ref{tab:datastat}.
\begin{itemize}
    \item \textbf{FB15k-237}~\cite{transe} is a subset of FreeBase~\cite{freebase} which contains facts extracted from Wikipedia. It consists of named entities and edges which determine the type of relation between the entities. The dataset is a benchmark for  evaluating KG embedding methods for link prediction.
    \item \textbf{ogbl-citation2}~\cite{ogbdatasets} is a citation graph extracted from Microsoft Academic Graph (MAG)~\cite{wang2020microsoft}.  Each vertex is associated with a 128-dimensional feature vector created using Word2Vec\cite{word2vec} on paper title and abstract. 
\end{itemize}

\begin{table}[h]
\centering
  \caption{Dataset statistics.}
  \label{tab:datastat}
  \begin{tabular}{cccc}
    \toprule
     Dataset & FB15k-237 & ogbl-citation2 \\
    \midrule
    \# Entities & 14,541 &  2,927,963\\
    \# Relations & 237 &  1   \\
    \# Features & - & 128  \\ 
    \# Train edges & 272,115  & 30,387,995 \\
    \# Valid edges & 17,535 &  86,596 \\
    \# Test edges & 20,466 &  86,596 \\
  \bottomrule
\end{tabular}
\end{table}

We use vertex-cut based partitioning (KaHIP~\cite{kahip, schlag2018scalable}) to obtain a disjoint subsets of training edges followed by neighborhood expansion to include all $n$-hop neighbors. Table~\ref{tab:part_stats} shows the statistics of the datasets after graph partitioning and neighborhood expansion. \textit{core edges} are the average number of edges in the partitions before expansion, and \textit{total edges} (sum of \textit{core edges} and \textit{support edges}) is the average number of edges after neighborhood expansion to 2 hops.
Replication Factor (RF) shows the quality of partitioned data after neighborhood expansion. It is calculated as:
\begin{equation}
\label{eq:rf}
    RF(P_1, P_2, ... , P_p) = \frac{1}{|V|}\sum_{i\in[p]}|V(E_i)|,
\end{equation}
where $P_i$ is the $i$th partition, $|V|$ is the total number of vertices in the graph,  $E_i$ is the edges in the $P_i$ partition, and $V()$ returns the set of vertices in a set of edges. The Table~\ref{tab:part_stats} shows that size of partitions highly depends on the size of the graph and number of hops($k$) for expansion. For smaller graph at FB15k-237 $k=2$, the partition size is almost equal to the number of original graph, and RF increases drastically as number of partitions increases. This trend is not seen  in ogbl-citation2 datasets  due to its large size. 

\begin{table}[!h]
\centering
\caption{Partition statistics (numbers in core edges and total edges columns are the average and standard deviation).}
\label{tab:part_stats}
\begin{tabular}{ccccc}
\toprule
\multicolumn{1}{c}{Dataset} & \# partitions & \#core edges & \#total edges  &RF\\
\midrule
\multirow{2}{*}{FB15k-237}  & 2  & 136k $\pm$ 4.5k & 270k $\pm$ 3.1k & 1.98  \\
                            & 4  & 68k $\pm$ 6.6 k & 265k $\pm$ 5.1k & 3.90 \\
                             & 8  & 34k $\pm$ 3.2k & 263k $\pm$ 4.0k & 7.75 \\
\midrule
\multirow{2}{*}{ogbl-citation2} & 2  &  15M $\pm$ 485K &    19M $\pm$ 95k & 1.45\\
                            & 4 & 7.5M $\pm$ 453K & 13M $\pm$ 1.2M & 2.26 \\
                             & 8  & 3.8M $\pm$ 189K  & 10M $\pm$ 1.1M &  3.58  \\
\bottomrule
\end{tabular}
\end{table}

\subsection{Hyperparameters}
We used a two layer RGCN~\cite{rgcn} model as a KG embedding method for link prediction and trained it in a distributed setting. For FB15k-237 dataset, we used the same hyperparameters as described in ~\cite{rgcn} except the embedding size. We found out that an embedding size of 75 dimensions produces comparable results with the original setting. As the dataset is small, we trained it using full edge batch.  For ogbl-citation2 dataset, we chose an embedding size of 32, learning rate 0.01, dropout 0.2, 1 negative sample per positive sample, and basis decomposition with two basis functions for regularization.
 For fairness of comparison, we used these hyperparameters in all training scenarios including non-distributed training. The number of partitions is equivalent to the number of distributed trainer processes. Since, we have two GPUs per compute node, we ran two trainers per machine. 

\subsection{Distributed Training Results}
In this section, we present the experimental results of our proposed approach. We compare our proposed distributed training with a non-distributed training setup (1 Trainer). The non-distributed training process trains on the full graph data.  We answer the following questions:
\begin{itemize}
    \item Does our distributed training approach have equivalence to non-distributed training (\textit{Accuracy})?
    \item Does our distributed training approach accelerate the training (\textit{Scalability} and \textit{Effect of number of model updates})?
    \item Does our partitioning strategy effectively balance the workload (\textit{Effect of partitioning})? 
\end{itemize}

\subsubsection{Accuracy}
\label{subsec:eval_accu}

\begin{table*}[t]
\centering
\caption{MRR, Hits@k and epoch time/speedup of RGCN non-distributed and distributed training on 3 datasets.}
\label{tab:results}
\begin{tabular}{cccccccc}
\toprule
 \multirow{2}{*}{\#Trainers} & \multicolumn{3}{c}{FB15k-237} & & \multicolumn{3}{c}{ogbl-citation2} \\
\cline{2-4}
\cline{6-8}
   & MRR  & Hits@1  & Ep. Time(s)/speedup & &MRR   & Hits@1  & Ep. Time(m)/speedup \\
\midrule
 1              &  0.22   &     0.138   & 5.09/-  &&   0.620  &  0.494  &  112/- \\
\midrule
2               &  0.22   &     0.136 & 4.03/1.25x &&  0.621   &   0.492 &  44/2.54 \\
4                &  0.21  &     0.130  &  3.62/1.40x &&  0.620   &   0.493  & 16/7x \\
8               &  0.21   &  0.124  & 3.54/1.43x  && 0.617 &     0.494  & 7/16x \\
\bottomrule
\end{tabular}
\end{table*}
We compare the performance of our distributed training approach with the non-distributed training setting. In case of distributed training, the number of trainers varies from 2 to 8 trainers with each compute node running two trainers.
Since, FB15k-237 is a small dataset, we performed full batch training. In case of ogbl-citation2, we performed mini-batch training, and the  mini-batch size is approximately 118k. For FB15k-237 dataset, we follow the filtered settings for evaluations. Since the number of edges for dataset ogbl-citation2 is very large, the dataset has provided 1000 candidate negative \textit{target} vertices for each test and validation edge for evaluation.
%The evaluation on these two datasets shows that our distributed training approach produces results similar to non-distributed training. 
We trained RGCN models on the two datasets, selected the best models and report the results on the test data. Ogbl-citation2 reached to maximum accuracy within 100 epochs. As shown in Table~\ref{tab:results}, the results (MRR and Hits@k) on benchmark dataset FB15k-237 shows that our distributed training approach produces similar results to the non-distributed setting which are also comparable to the numbers reported in the original RGCN paper. For the larger dataset, ogbl-citation2, we observe a similar trend, i.e our distributed training approach produces results similar to non-distributed training. The results also verify that our constraint-based negative sampling strategy is effective in training as no deterioration of the scoring metrics is observed.

\subsubsection{Scalability}
\begin{figure}[!h]
\centering
    \includegraphics[width=\linewidth]{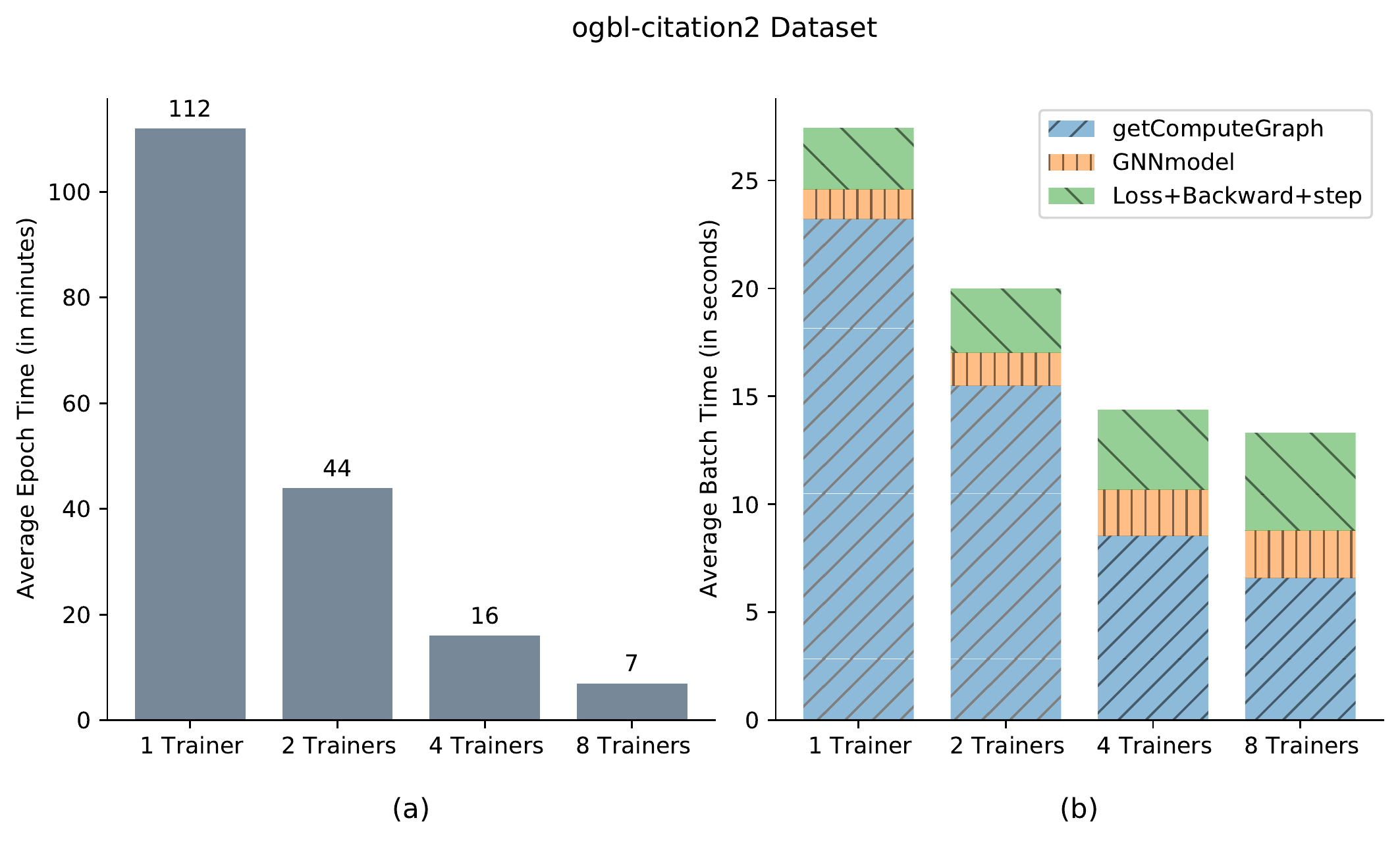}
    \caption{(a) Average running time per epoch; (b) average running time of components in batch for ogbl-citation2 dataset (same batch size across all trainers produces 256, 128, 64, 32 batches for 1,2,4,8 trainers respectively).}
\label{fig:batch_time}
\end{figure}

We used the same settings as described in Section~\ref{subsec:eval_accu} for evaluating the scalability of our approach. We did not apply any sampling strategy (vertex drop or edge drop) during training. As shown in Table~\ref{tab:results}, the speed up for FB15k-237 dataset is lower than linear. It is because the size of the partitions are equal to the original dataset as shown in Table~\ref{tab:part_stats}. For ogbl-citation2 dataset, we achieved a speedup of approximately 16x with 8 trainers compared to 1 trainer. 
We performed an in-depth analysis of the running times of the major computational components described in lines 4-10 of Algorithm~\ref{algo:train} (\textit{getComputeGraph, GNNmodel, loss+backward+step}) in order to quantify the contributions of these components to the overall speedup. Figure~\ref{fig:batch_time}(a,b) shows the average epoch time and average running time of different components in a batch. The component, \textit{getComputeGraph} is a very compute intensive operation as it depends on the partition size. The function returns a computational graph for an edge mini-batch. It is essential function as the it enables us to train on large graphs.  
The running time of this operation decreases as we increase the number of trainers from 1  to 8, because the size of the partitions decreases. In the case of 8 partitions for ogbl-citation2, the size of each partition decreases by one-third with respect to the full graph.
\textit{GNNmodel} produces the embeddings of the vertices in an edge mini-batch. Its running time in case of multiple trainers is slightly higher than for 1 trainer, because we run 2 trainers per machine which share the same resources. The running time of the third block of operations (\textit{loss+backward+step})  increases as we increase the number of trainers, because of the increase in communication cost for gradient sharing. 
The overall impact of these components  on training time varies because the number of batches (forward pass and backward pass) per epoch decreases from 256  to 32 for 1 trainer and 8 trainers respectively. 

\subsubsection{Convergence}

\begin{figure}
\centering
    \includegraphics[width=\linewidth]{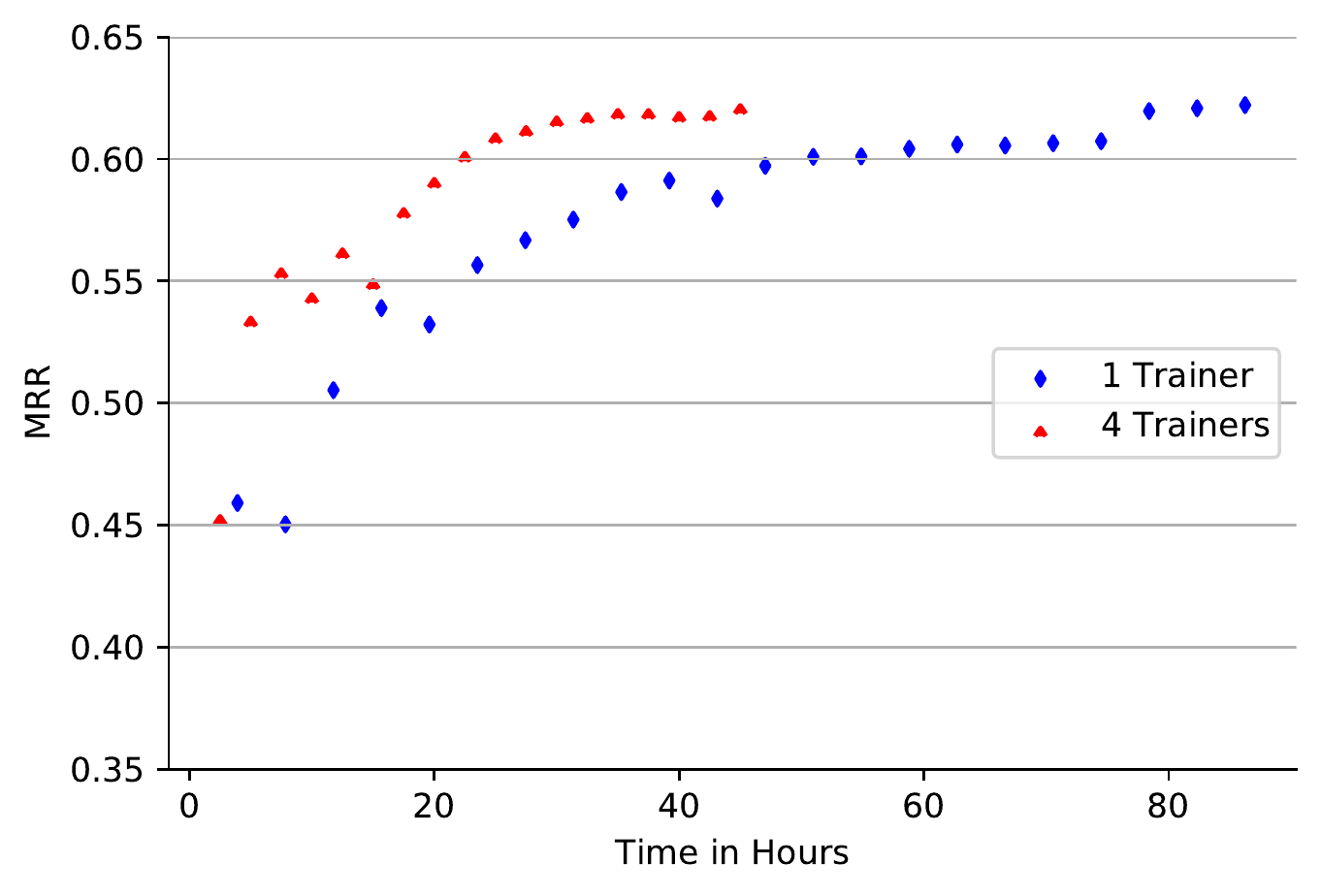}
    \caption{Convergence of distributed training on ogbl-citation2 dataset.}
\label{fig:conv_time}
\end{figure}

To show the convergence of our training approach, we used 4 trainers in a distributed setting.
Figure~\ref{fig:conv_time} shows that our distributed training converges faster and attains the same peak MRR score achieved by non-distributed training which takes a lot more time to converge. 

\subsubsection{Effect of number of model updates}
We study the effect of number of model updates on scalability. For this experiment, we fixed the number of batches in each epoch which in turn leads to a smaller batch size when going from 2 trainers to 8 trainers. Table~\ref{tab:num_batch} shows a speedup gain of 3.7x is achieved in ogbl-citation2 dataset. This is due to the fact that the overall impact of training components on running time is nearly equal as the number of forward and backward passes are same.

\begin{table}[h]
\centering
\caption{Epoch time of RGCN on ogbl-citation2 dataset with fixed \#model updates.}
\label{tab:num_batch}
\begin{tabular}{cccc}
\toprule
\#Vertices & \#Partitions & Time(hrs) & Avg. \#edges per batch\\
\midrule
1       & 1             &   1.96   & 118k\\
\midrule
1       & 2             &   1.30   &    60k \\
2       & 4             &   0.78   & 30k  \\
4       & 8             &   0.52  & 15k\\
\bottomrule
\end{tabular}
\end{table}

\subsubsection{Effect of Partitioning}

We examine the effectiveness of different partitioning strategies other than vertex-cut partitioning and their effect on running times. We used Random partitioning and METIS partitioning for this study. In Random partitioning, we randomly divide the edges into 4 partitions and then subsequently applied neighborhood expansion. In case of  METIS, the first hop neighbors of vertices are the core edges of a partition.  We applied neighborhood expansion on the partitions of vertices to include all $n$-hop neighbors of vertices in each partition. Table~\ref{tab:part_rm} shows the statistics of the partitions produced using these two methods. The size of partitions produced by Random partitioning strategy are equal to the original graph, whereas in case of METIS partitions, the size of the partitions is bigger than the partitions produced using vertex cut partitioning.

After obtaining the partitions, we trained the model with the same hyperparameters for the Random and Metis partitions, and measured the epoch times. In this experiment, we fixed the number of batches to 256 to have the same number of model updates in all different partitioning strategies. As shown in Table~\ref{tab:part_stats}, the epoch time in the case of Random partitioning is equal to the non-distributed training. This is because the size of each partition is equal to the full graph.  The epoch time for METIS partitions is 1.3x slower than vertex-cut based partitioning strategy (KaHIP). Hence, our partitioning strategy is aptly suited for link prediction task. 

\begin{table}[h]
\centering
\caption{Partition statistics of 4 partitions using METIS and random partitioning; Time per epoch(in hours) on ogbl-citation2 dataset with 2 nodes having 2 trainers each.}
\label{tab:part_rm}
\begin{tabular}{cccc}
\toprule
Partitioning & \#core edges & \#total edges & Time per epoch\\
\midrule
KaHIP+NE &4.1M $\pm$ 187K   & 12M $\pm$ 148k & 0.78\\
Metis+NE & 7.5M $\pm$ 3.0M& 16M $\pm$ 223K & 0.95 \\
Random+NE & 7.5M $\pm$ 0 & 29M $\pm$ 1k & 1.88\\
\bottomrule
\end{tabular}
\end{table}

\section{Related Work}
There has been a significant focus on developing frameworks~\cite{dgl, distdgl, openke,pbg, dglke, neugraph, distgnn, aligraph, euler} for graph embeddings that provide various functionalities and optimizations for training existing models, and implementing new models in an efficient and scalable manner. These frameworks support training on mutli-gpu multi-node clusters. The distributed frameworks partition the input graph, and each compute node runs a replica of the model to train on a partition. Aligraph~\cite{aligraph} provides a generic architecture on which various GNNs can be implemented. It consists of various components such as distributed graph storage and a caching strategy. Neugraph~\cite{neugraph} uses a vertex-program, graph-parallel abstraction to express GNN algorithms. NeuGraph performs GNN training on a a single node, multi-gpu system.
There are additional frameworks that focus on knowledge graph embeddings such as OpenKE~\cite{openke}, PyTorch-BigGraph(PBG)~\cite{pbg}, and DGL-KE~\cite{dglke}. OpenKE\cite{openke} is designed as a standalone system for knowledge graph embeddings, and does not scale to very large graphs. Both DGL-KE~\cite{dglke} and PBG~\cite{pbg} are distributed frameworks for knowledge graph embedding models. However, all these three frameworks are designed for embedding models which are either translation-based or semantic-based such as TransE~\cite{transe} or DistMult~\cite{distmult}. When computing the embedding of entities, these models do not require information from neighbors as required in GNNs. Thus, these frameworks cannot be used to train GNN based knowledge graph embedding models~\cite{rgcn, relatt, relgnn} where nodes have dependencies up to $k$-hop neighbors.

DistDGL~\cite{distdgl} is a distributed framework for GNN training based on DGL~\cite{dgl}. It employs a mini-batch training approach to train large graphs, besides various other optimizations.  DistDGL distributes the graph by partitioning it using Metis~\cite{metis}, and the graph partitions along with the associated data are assigned compute nodes for training. The vertex data is stored in a Key-Value store and fetched during training. For transductive models, DistDGL provides a distributed sparse embedding layer and updates it using sparse updates during training. DistDGL has shown superior performance in scaling models for node classification. However, as discussed in previous section, the edge cut partitioning based methods are not suitable for link prediction.

\section{Conclusion}
We proposed algorithmic approaches for distributed training of GNN-based knowledge graph embedding models. Our approach is agnostic to the used knowledge graph embedding model. We used a vertex cut partitioning approach along with neighborhood expansion method to make the partitions self-sufficient such that no data is transferred across partitions during training. We introduced edge mini-batch training for large partitions that enables us to train on large partitions with limited system memory.  Moreover, we applied constraint-based negative sampling to exploit the local partitions to generate the negative samples for training. Our experimental evaluation shows a super linear speedup on a cluster of machines without sacrificing model accuracy, and our approach converges faster than non-distributing training. 

\bibliographystyle{ACM-Reference-Format}
\bibliography{references}

\end{document}